\documentclass[letterpaper, 10 pt, conference]{ieeeconf}  

\IEEEoverridecommandlockouts                              

\usepackage{cite}

\usepackage{makecell}

\usepackage{float} 
\usepackage{graphicx}
\usepackage{booktabs}
\usepackage{multirow}
\usepackage{xcolor}
\usepackage{caption}
\usepackage{amsmath,amssymb,amsfonts}
\usepackage{algorithmic}
\usepackage{graphicx}
\usepackage{textcomp}
\usepackage{xcolor}
\usepackage{bm}
\def\BibTeX{{\rm B\kern-.05em{\sc i\kern-.025em b}\kern-.08em
    T\kern-.1667em\lower.7ex\hbox{E}\kern-.125emX}}
\begin{document}

\newcommand{\rt}{\textcolor[rgb]{1,0,0}}
\newcommand{\bt}{\textcolor[rgb]{0,0,1}}
\newcommand{\pt}{\textcolor[rgb]{0.5,0,0.5}}

\title{\LARGE \bf
LoGS: Visual Localization via Gaussian Splatting\\ with Fewer Training Images
}

\author{
    Yuzhou Cheng\textsuperscript{1}, Jianhao Jiao\textsuperscript{1*}, Yue Wang\textsuperscript{2}, and Dimitrios Kanoulas\textsuperscript{1, 3}
    \thanks{\textsuperscript{1}Robot Perception and Learning Lab, Intelligent Robotics, Department of Computer Science, University College London, Gower Street, WC1E 6BT, London, UK. {\tt\small \{yuzhou.cheng.23, ucacjji, d.kanoulas\}@ucl.ac.uk}}
    \thanks{\textsuperscript{2}Zhejiang University, Hangzhou, Zhejiang, China. {\tt\small wangyue@iipc.zju.edu.cn}}
    \thanks{\textsuperscript{3}AI Centre, Department of Computer Science, University College London, Gower Street, WC1E 6BT, London, UK and Archimedes/Athena RC, Greece.}
    \thanks{This work was supported by the UKRI FLF [MR/V025333/1] (RoboHike).  For the purpose of Open Access, the author has applied a CC BY public copyright license to any Author Accepted Manuscript version arising from this submission.}
    \thanks{\textsuperscript{*}Corresponding Author: Jianhao Jiao}}

\maketitle

\begin{abstract}
Visual localization involves estimating a query image’s 6-DoF (degrees of freedom) camera pose, which is a fundamental component in various computer vision and robotic tasks. This paper presents LoGS, a vision-based localization pipeline utilizing the 3D Gaussian Splatting (GS) technique as scene representation. This novel representation allows high-quality novel view synthesis. During the mapping phase, structure-from-motion (SfM) is applied first, followed by the generation of a GS map. During localization, the initial position is obtained through image retrieval, local feature matching coupled with a PnP solver, and then a high-precision pose is achieved through the analysis-by-synthesis manner on the GS map. Experimental results on four large-scale datasets demonstrate the proposed approach’s SoTA accuracy in estimating camera poses and robustness under challenging few-shot conditions.
\end{abstract}

\section{Introduction}
\subsection{Motivation}
In an increasingly automated world, the ability of robots to understand and navigate their surrounding environment has become crucial for numerous applications, ranging from autonomous vehicles and extended reality (XR) to industrial automation and disaster response. Visual localization is at the core of capabilities, allowing robots to accurately determine their six degrees of freedom (6-DoF) position and orientation. 

Current visual localization methods fall into three major types: absolute pose regression (APR) \cite{kendall2015posenet,clark2017vidloc,geometricloss2017,directposenet2021,multiscene2021,chen2024map}, structure-based \cite{sattler2016efficient, sarlin2020superglue, sarlin2019coarse, shotton2013scene,brachmann2017dsac,brachmann2021visual,li2020hierarchical,huang2021vs,yang2019sanet}, and analysis-by-synthesis \cite{yen2021inerf, lin2023parallel,trivigno2024unreasonable, chen2024neural, zhou2024nerfect,sun2023icomma,bortolon20246dgs} methods. 
\textbf{APR} estimates the camera pose directly from neural networks but need help with accuracy and generalization. 
\textbf{Structure-based} approaches contain feature matching-based (FM) \cite{sattler2016efficient, sarlin2020superglue, sarlin2019coarse} and scene coordinate regression (SCR) \cite{shotton2013scene,brachmann2017dsac,brachmann2021visual,li2020hierarchical,huang2021vs,yang2019sanet}. FM identifies 2D-3D correspondences between image projections and spatial coordinates in the point cloud, while SCR directly predicts such correspondences each pixel through a trained neural network. 
Typically, a geometric solver such as the PnP-RANSAC estimates the camera poses these 2D-3D correspondences. 
FM pipelines have been widely adopted, but their accuracy is usually lower than that of SCR if the model is trained with sufficient data. 
Nevertheless, many SCR networks are specifically designed for localization, making them an additional burden for robots.

Recently, iNeRF \cite{yen2021inerf, lin2023parallel} emerged as an \textbf{analysis-by-synthesis} approach that iteratively inverts neural radiance fields (NeRFs) to align camera poses. Nonetheless, these approaches suffer from time limitations due to low rendering speed. 3D Gaussian Splatting (GS) \cite{kerbl20233d}, a paradigm-shifting Novel View Synthesis technique, achieves comparable render quality and real-time rendering. It rasterizes a collection of Gaussian ellipsoids to approximate a scene’s appearance. Analysis-by-synthesis localization using 3DGS as the map representation \cite{sun2023icomma, bortolon20246dgs} has started to gain attention. They have yet to be tested on large-scale datasets \cite{shotton2013scene, kendall2015posenet, valentin2016learning} and lack comparisons to baselines in other categories.

\subsection{Contributions}
This paper introduces a novel visual localization pipeline, termed LoGS, which employs GS as the foundational map structure. 
Especially, LoGS addresses challenges related to data scalability. 
As we want: ``You don't need a lot to make a difference.'' 
Training a environmental representation with only dozens or even just a few images generally alleviates data scarcity and reduces resource requirements, but at the cost of accuracy decay \cite{dong2022visual}. 
This few-shot setting \cite{dong2022visual} tests a pipeline's robustness and generality as well, where many of the aforementioned neural network-based methods tend to fail. 
Our method, on the contrary, outperforms the state-of-the-art (SoTA) using only 0.5\% to 1\% of the training images. For example, by utilizing only 20 out of 4000 images, we achieve a median translation error of 0.5 cm and a median rotation error of 0.16° (see TABLE \ref{7scenes_table2}) in the CHESS scene from the 7-scenes dataset \cite{shotton2013scene}. 
This is crucial for practical applications that require rapid deployment.

We obtain a point cloud for GS map initialization by performing Structure-from-Motion (SfM) with advanced feature-matching.
Then, we utilize depth clues and regularization strategies to build a high-resolution GS map. LoGS estimates a rough pose through PnP-RANSAC on the SfM point cloud when localization starts. LoGS then minimizes the photometric loss between the query image and the rendered images on the GS map to obtain an exceptionally accurate final pose. We also propose masking policies to choose the most representative pixels for residual comparison. Our pipeline achieves SoTA accuracy across four large-scale localization benchmarks \cite{barron2022mip, mildenhall2019local,shotton2013scene, kendall2015posenet} covering indoor and outdoor environments. 
In summary, the contributions of this work are threefold:
\begin{itemize}
\item We present a novel visual localization pipeline with 3DGS as the core map representation, which operates in a hierarchical manner.

\item  Extensive experiments have been conducted on four real-world full/few-shot benchmarks. LoGS is on par with or sets new baselines for these datasets.

\item We demonstrate the practical effectiveness of adding depth clues and regularization strategies for GS map formation and the usefulness of adding different masks for photometric residuals' comparison.  
\end{itemize}

\section{Related Work}

\subsection{Absolute Pose Regression}
Absolute Pose Regression (APR) entails training neural networks to directly regress a 6-DoF camera pose from an image. PoseNet \cite{kendall2015posenet} marks the first APR method utilizing a CNN framework. Following PoseNet, enhancements such as temporal information \cite{clark2017vidloc}, geometric loss function \cite{geometricloss2017}, and photometric consistency \cite{directposenet2021} have refined the accuracy of APR. Applying Transformer mechanisms in Multi-Scene APR \cite{multiscene2021} and map-relative pose regression \cite{chen2024map} have also propelled the field. However, APR still suffers from precision and generality \cite{sattler2019understanding, liu2024hr}. Our LoGS overcomes the significant drop in accuracy prediction experienced by these direct methods on few-shot training images.

\subsection{Structure-based Localization}

Structure-based localization involves: 1) identifying correspondences between 2D image pixels and 3D scene points and 2) solving for the camera pose through a geometric solver such as PnP-RANSAC. 
Traditional FM approaches \cite{sattler2016efficient, sarlin2020superglue, sarlin2019coarse} establish correspondences via 2D-2D feature matching while recent Scene Coordinate Regression (SCR) methods regress pixels to 3D coordinates. Pioneering SCR uses Regression Forests \cite{shotton2013scene} for RGB-D camera localization. Neural network approaches have gradually outperformed Regression Forests in recent years. DSAC \cite{brachmann2017dsac}\cite{brachmann2021visual} employs CNNs to predict scene coordinates and score hypotheses, introducing a differentiable RANSAC algorithm. Region classification \cite{li2020hierarchical} and the segmentation branch \cite{huang2021vs} are later introduced to enhance scene understanding. There is also attention on scene-agnostic coordinates regression \cite{yang2019sanet} where model parameters and scenes are independent. Extending FM's pipeline, LoGS further improves the accuracy and achieves SoTA results on the 7-scenes and Cambridge Landmarks datasets through an additional refinement step.

\subsection{Analysis-by-synthesis}
Analysis-by-synthesis methods optimize the camera pose by reducing the \(L_1\) or \(L_2\) norm of the difference in pixel-level features between a synthesized image and the query image. They either independently achieve relocalization or serve as pose refinement modules. This approach has its roots in many visual tracking components \cite{newcombe2011dtam, engel2014lsd,engel2017direct}. 
These works \cite{yen2021inerf, lin2023parallel, zhou2024nerfect} use iNeRF to refine camera poses through photometric loss while \cite{trivigno2024unreasonable, chen2024neural} render and align higher dimensional features for each pixel. Recent analysis-by-synthesis methods use 3D GS as the scene representation, as seen in works \cite{sun2023icomma}\cite{bortolon20246dgs}. Nevertheless, pose refinements of many methods mentioned above are prone to converge to local optima. To tackle this issue, LoGS builds a fine-grained GS map with depth clues and regularization strategies and designs masks that filter pixels to avoid local convergence.

\section{Preliminaries}
The 3D GS \cite{kerbl20233d} is built upon three components \cite{chen2024survey}. The first component is the basic scene representation. 3D Gaussians in space are colored ellipsoids whose transparency gradually decays from its center point according to a Gaussian distribution. The second component focuses on optimizing the properties of the 3D Gaussians. During optimization, 3DGS adopts an adaptive density control—adds and removes 3D ellipsoids to produce a compact and unstructured scene representation, typically resulting in several million Gaussians for a target scene. The last element of 3DGS is a rapid rendering strategy that leverages tile-based rasterization. 

We define a 3D Gaussian by its opacity \(\alpha\), color \(c\), center position \(\bm{\mu}\),  and 3D covariance matrix \(\mathbf{\Sigma}\):
\begin{align}
    G(\mathbf{x}) = [\alpha,c] e^{-\frac{1}{2} (\mathbf{x} - \bm{\mu})^{\top} \bm{\Sigma}^{-1} (\mathbf{x} - \bm{\mu})} .
\end{align}

3D Gaussians are projected into the 2D image plane for rendering. The resulting 2D covariance matrix \(\bm{\Sigma}'\) is derived from the viewing transformation \(\mathbf{W}\) and the 3D covariance matrix \(\bm{\Sigma}\):
\begin{align}
    \bm{\Sigma}' = \mathbf{JW}\bm{\Sigma} \mathbf{W}^\top \mathbf{J}^\top .
\end{align}

For a given pixel position $\mathbf{p}$, the distances to all overlapping Gaussians generate a sorted list of Gaussians \(\mathcal{N}\) \cite{chen2024survey}. Alpha compositing is applied to determine the pixel's color:
\begin{equation}
    C(\mathbf{p}) = \sum_{i=1}^{|\mathcal{N}|} c_i \alpha'_i \prod_{j=1}^{i-1} (1 - \alpha'_j) ,
\end{equation}
where \(c_i\) is the color after training. The opacity \(\alpha'_i\) is the multiplication outcome of the trained opacity \(\alpha_i\) and projected position within the Gaussian: \(
    \alpha'_i = \alpha_i \cdot \exp\left(-\frac{1}{2}(\mathbf{x}' - \bm{\mu}'_i)^\top {\bm{\Sigma}'}_i^{-1} (\mathbf{x}' - \bm{\mu}'_i)\right),
\) where \(\mathbf{x}'\) and \(\bm{\mu}'_n\) are coordinates in the projected space.
\begin{figure*}[t]
\centering
    \begin{minipage}{0.9999\textwidth}
    \includegraphics[width=\textwidth]{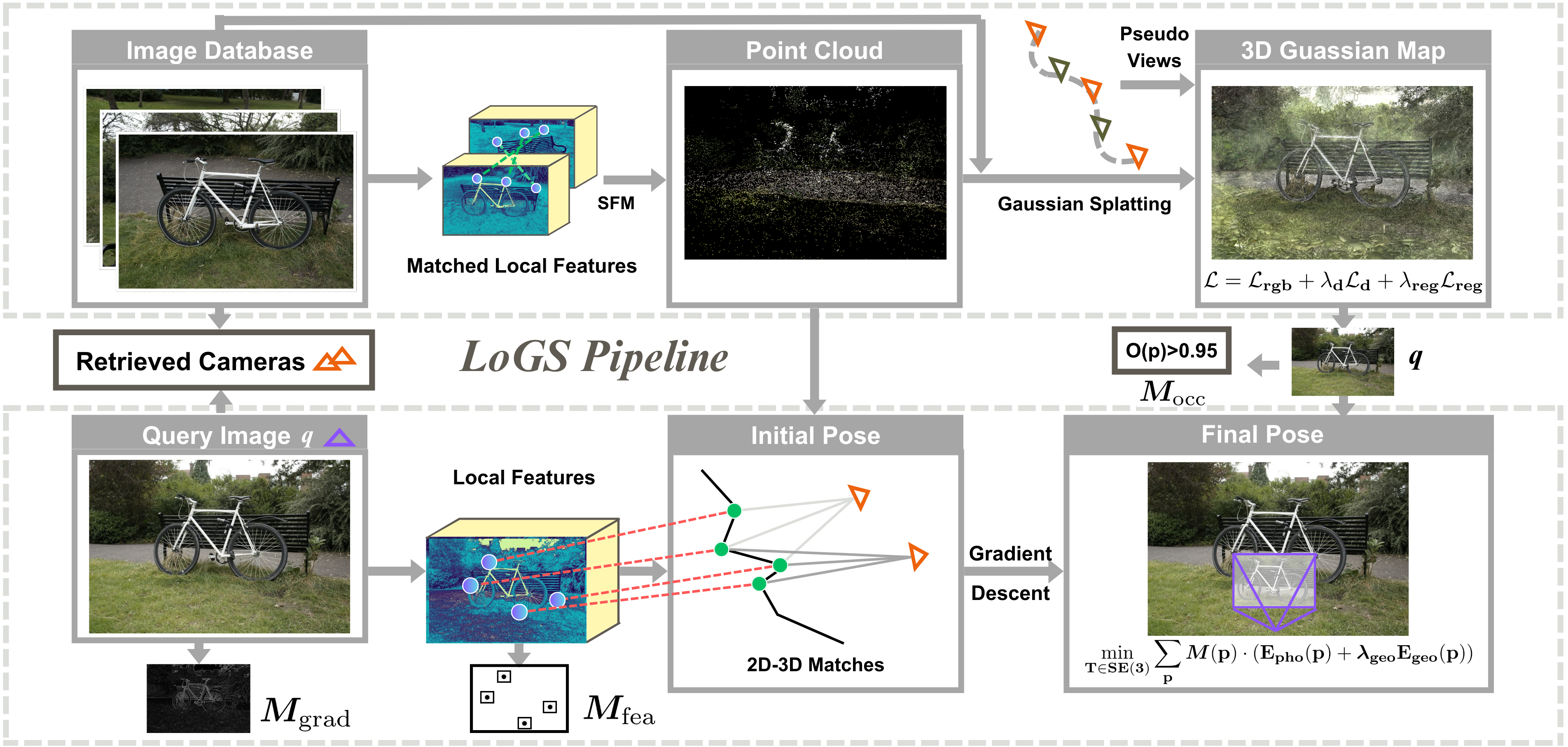}
    \end{minipage}
\caption{An illustration of the LoGS pipeline where the localization process aligns with the mapping. }
\label{pipeline}
\end{figure*}
By accumulating the distance along the ray, we can also define a differentiable rendered depth:
\begin{align}
D(\mathbf{p}) &= \sum_{i=1}^{|\mathcal{N}|} d_i \alpha_i' \prod_{j=1}^{i-1} \left(1 - \alpha_i'\right),
\end{align}
which can be compared with the input depth map if the query image has a depth channel.
Additionally, we render an occupancy image to determine visibility:
\begin{align}
O(\mathbf{p}) &= \sum_{i=1}^{|\mathcal{N}|} \alpha_i' \prod_{j=1}^{i-1} \left(1 - \alpha_i'\right).
\end{align}
This occupancy image measures the confidence level of the Gaussian ellipsoids' contribution at a given pixel. 

\section{Methodology}
\subsection{Mapping}
\textbf{SfM}: A random initial distribution can result in some Gaussians being unable to be optimized to their ideally optimal positions, leading to artifacts such as floaters. This, in turn, can affect the final image rendering quality. If we already have an SfM point cloud distribution at the start of GS map construction, we can initialize an ellipsoid at each point, which gives a relatively good representation from the beginning. Thus, LoGS first utilizes SuperPoint \cite{detone2018superpoint} and SuperGlue \cite{sarlin2020superglue} to extract features and perform feature matching on the images in the database. Then, an accurate sparse point cloud is constructed through SfM triangulation.\\

\textbf{GS map}: Given all the renders, we design a loss function to optimize learnable parameters in the GS map. We first reduce the photometric residual:
\begin{align}
\mathcal{L}_{rgb} &= \left\lVert C - \bar{C} \right\rVert_1 ,
\end{align}
where \( C \) is the rendered color image from the Gaussians and ground truth pose \( T \), and \( \bar{C} \) is the ground truth color image. 

When the images for training have a depth channel, we similarly express the geometric loss using the $L_1$ norm:
\begin{align}
\mathcal{L}_{d} &= \left\lVert D - \bar{D} \right\rVert_1 ,
\end{align}
where \( D \) is the rendered depth image and \( \bar{D} \) is the ground truth depth image.

When ground truth depth is absent, we generate monocular depth maps \(\hat{D}\) for training images using the pre-trained Dense Prediction Transformer (DPT) \cite{ranftl2021dpt} to regularize the training. We apply a relaxed relative loss using Pearson correlation \cite{zhu2023fsgs} between the estimated depth $\hat{D}$ and rendered depth \( D \). This method measures the distributional differences between the two depth maps:
\begin{equation}
\mathcal{L}_{reg}(D, \hat{D}) = \frac{\text{Cov}(D, \hat{D})}{\sqrt{\text{Var}(D)\text{Var}(\hat{D})}}.
\end{equation}
Experimental results show that adding this regularization term improves the quality of novel view synthesis even when the dataset includes ground truth RGB-D images. This improvement is due to the continuity of pseudo depths, which filter out isolated artifact Gaussians. 

In sum, we reach the following optimization objective for each training image in the image database:
\begin{align}
\mathcal{L} = \mathcal{L}_{rgb} + \lambda_{d} \mathcal{L}_{d} + \lambda_{reg} \mathcal{L}_{reg}.
\end{align}

When there are very few training images, the coverage of the scene is incomplete, and over-fitting happens. LoGS applies the $\mathcal{L}_{reg}$ loss on pseudo-views. Given a set of $N$ train-view poses $\{T_1, \dots, T_N\}$, where each pose $T_{i} = \{R_i, t_i\}$, we find a permutation $\pi$ that minimizes the total pairwise distance:
\begin{align}
\min_{\pi} \sum_{i=1}^{N-1} d(T_{\pi(i)}, T_{\pi(i+1)}),
\end{align}
where $ d(T_i, T_j) = \left\lVert t_i - t_j \right\rVert_2 $ is the \(L_2\) translation error. 

$K$ pseudo views are interpolated between consecutive poses $T_{\pi(i)}$ and $T_{\pi(i+1)}$ with Spherical Linear Interpolation (SLERP):
\begin{align}
\left\{
\begin{aligned}
t^{(k)} &= (1 - \alpha(k)) t_i + \alpha(k) t_{i+1} \\
R^{(k)} &= \text{SLERP}(R_i, R_{i+1}, \alpha(k))
\end{aligned}
,\right.
\end{align}
where \(
\alpha(k) = \frac{1 - \cos\left(\frac{k\pi}{K+1}\right)}{2}
\). A series of smoothly transitioning pseudo views are thus generated between real views.

\subsection{Localization}

\textbf{Initial Pose Estimation}:
To estimate an initial pose, we employ a feature matching-based approach \cite{sarlin2018leveraging} on the SfM point cloud, which consists of prior retrieval, covisibility clustering, local matching, and localization.

1) Prior Retrieval: Compare the query image with database images using global descriptors from NetVLAD \cite{arandjelovic2016netvlad}. The k-nearest neighbors represent potential locations within the map. 

2) Covisibility Clustering: cluster the neighbors based on the covisibility of 3D structures—two frames belong to the same place if they observe common 3D points. Then, we perform independent local searches in each place.

3) Local Matching and Localization: Beginning with the place that contains the most number of nearest neighbors, we traverse through every place. We obtain the geometric relationship by matching local descriptors between the 3D points in the place and critical points in the query image with SuperGlue \cite{sarlin2020superglue}. Finally, we check the geometry consistency and estimate a pose by solving the PnP-RANSAC problem.\\

\textbf{Iterative Pose Refinement}:
Given limitations in matching accuracy, point cloud precision, and the visibility overlap of retrieved images, the initial pose is partially accurate. However, it is a strong starting point for further pose refinement. 3DGS enables a direct, nearly linear (projective) gradient flow between the parameters and the rendered output. As a result, we refine the pose through iterative updates using gradient-based optimization, taking advantage of differentiable rendering for both RGB and depth: 
\begin{align}
    \hat{T} = \arg\min_{T \in \text{SE}(3)} \mathcal{L}(T \mid \mathcal{I}, \mathcal{G}) ,
\end{align}
where $\mathcal{I}$ is the query image and $\mathcal{G}$ is the GS map.

At each iteration, we optimize and update the camera pose with respect to the GS map. In the monocular case, we minimize the following photometric residual \cite{matsuki2024gaussian}:
\begin{align}
E_{pho}(\mathbf{p}) &= | (e^{a}\cdot C(\mathbf{p}) + b)- \bar{C}(\mathbf{p}) | ,
\end{align}

where \( C(\mathbf{p}) \) is the color of the rendered image at pixel $\mathbf{p}$, and \( \bar{C}(\mathbf{p}) \) is the color of the observed image at the same pixel position. We optimize affine brightness parameters $a$ and $b$ for varying exposure. These two parameters are vital for controlling illumination changes, especially in outdoor environments.

When a depth channel exits, we similarly define the depth residual between rendered depth and ground truth depth at a given pixel $\mathbf{p}$:
\begin{align}
E_{geo}(\mathbf{p}) &= |D(\mathbf{p}) - \bar{D}(\mathbf{p})| .
\end{align}

To mitigate the impact of noise in the scene representation that could distort the rendered images, we carefully designed a mask to select only information-rich pixels for comparison. This filter results in a more robust objective function and prevents the optimizer from sinking into local optima.

An edge detector is used to select pixels above a certain threshold, capturing important structural information in the image and reducing the amount of data that needs to be processed. The gradient mask \( M_\text{grad} \) is then defined as: 
\begin{align}
M_\text{grad}(\mathbf{p}) =
\begin{cases}
1, & \text{if } |\nabla(\mathbf{p})| > \tau_\text{grad}, \\
0, & \text{otherwise}.
\end{cases}
\end{align}
where $
 |\nabla(\mathbf{p})|$ is the gradient magnitude of the Scharr operator.

We used SuperPoint \cite{detone2018superpoint} during SfM to extract local descriptors and key points. These points effectively identify corners or blobs in the image. Around each significant feature point, we select a small area as the region of interest:
\begin{align}
M_\text{fea}(x, y) =
\begin{cases}
1, & \text{if } \exists (x_i, y_i) ~s.t.~ |x - x_i| \leq \tau_\text{fea} \\
   & \text{and } |y - y_i| \leq \tau_\text{fea}, \\
0, & \text{otherwise}.
\end{cases}
\end{align}

Opacity mask $M_\text{occ}$ focuses on pixels that contain Gaussian ellipsoid information rather than on arbitrary pixels:
\begin{align}
M_\text{occ}(\mathbf{p}) =
\begin{cases}
1, & \text{if } O(\mathbf{p}) > \tau_\text{occ}, \\
0, & \text{otherwise}.
\end{cases}
\end{align}

To summarize, we reach the following optimization objective for the pose:
\begin{align}
\min_{T \in SE(3)} \sum_{\mathbf{p}} M(\mathbf{p})\cdot (E_{pho}(\mathbf{p}) + \lambda_{geo} E_{geo}(\mathbf{p})) ,
\end{align}
where $
M = (M_{\text{grad}} \cup M_{\text{fea}}) \cap M_{\text{occ}}$.

\section{Experiments}
\subsection{Datasets}
We choose the Mip-NeRF 360 \cite{barron2022mip} and LLFF \cite{mildenhall2019local} datasets to compare analysis-by-synthesis baselines \cite{yen2021inerf,sun2023icomma}. The Mip-NeRF 360 dataset consists of nine scenes, five outdoors and four indoors, while the LLFF has complex real-world scenes for rendering novel views. To compare with other mainstream localization methods, we choose the widely-used indoor 7-scenes dataset \cite{shotton2013scene} and the outdoor Cambridge Landmarks dataset \cite{kendall2015posenet}.

\subsection{Metrics}
Translation error is the norm of the difference between the ground truth pose’s position and the
estimated pose’s position, while the rotation error is the angle between the ground truth orientation and the estimated orientation. \textbf{Success rate} corresponds to the proportion of rotation error less than a threshold (5 degrees) and the proportion of translation error less than a threshold (5 cm) \cite{brachmann2017dsac, li2020hierarchical, dong2022visual}. \textbf{Median pose error} refers to the median of the translation errors and the median of the rotation errors among all testing images. 

\begin{table*}[!t]
\centering
\caption{Quantitative comparison of methods on the 7-Scenes dataset with DSLAM ground truth. Results: AS \cite{sattler2016efficient}, HLoc \cite{sarlin2020superglue, sarlin2019coarse}, HSCNet \cite{li2020hierarchical}, DSAC* \cite{brachmann2021visual}, SP+Reg \cite{dong2022visual}, FSRC \cite{dong2022visual}. Fewshot results are from \cite{dong2022visual}.}
\renewcommand{\arraystretch}{1.5}
\resizebox{\textwidth}{!}{
\begin{tabular}{l|c|c|c|c|c|c|c||c|c|c|c|c|c|c}
\toprule
\multirow{1}{*}{\makecell{Methods }} & \multirow{2}{*}{\#Images} & \multicolumn{6}{c||}{Original training (median pose error in cm/°)} & \multirow{2}{*}{\#Images} & \multicolumn{6}{c}{Few-shot training (median pose error in cm/°)} \\ \cline{3-8} \cline{10-15}

 (DSLAM GT)&  & AS  & HLoc  & HSCNet & DSAC* & ACE & Ours  &  & HLoc & DSAC* & HSCNet & SP+Reg & FSRC & Ours \\ \hline
CHESS & 4000 & 3/0.87 & \textcolor{red}{2}/0.85 & \textcolor{red}{2}/\textcolor{blue}{0.7} & \textcolor{red}{2}/1.10 & \textcolor{red}{2}/\textcolor{blue}{0.7} & \textcolor{red}{2.0}/\textcolor{red}{0.62} & 20 & \textcolor{blue}{4}/1.42 & \textcolor{red}{3}/\textcolor{blue}{1.16} & \textcolor{blue}{4}/1.42 & \textcolor{blue}{4}/1.28 & \textcolor{blue}{4}/1.23 & \textcolor{red}{3}/\textcolor{red}{1.00} \\ \hline

FIRE & 2000 & \textcolor{red}{2}/1.01 & \textcolor{red}{2}/\textcolor{blue}{0.94} & \textcolor{red}{2}/\textcolor{blue}{0.9} & \textcolor{red}{2}/1.24 & \textcolor{red}{2}/\textcolor{blue}{0.9} & \textcolor{red}{1.8}/\textcolor{red}{0.70} & 10 & \textcolor{blue}{4}/1.72 & 5/1.86 & 5/1.67 & 5/1.95 & \textcolor{blue}{4}/\textcolor{blue}{1.53} & \textcolor{red}{2}/\textcolor{red}{0.90} \\ \hline

HEADS & 1000 & \textcolor{red}{1}/0.82 & \textcolor{red}{1}/\textcolor{blue}{0.75} & \textcolor{red}{1}/0.9 & \textcolor{red}{1}/1.82 & \textcolor{red}{1}/\textcolor{red}{0.6} & \textcolor{red}{1.0}/\textcolor{red}{0.64} & 10 & 4/1.59 & 4/2.71 & \textcolor{blue}{3}/1.76 & \textcolor{blue}{3}/2.05 & \textcolor{red}{2}/\textcolor{blue}{1.56} & \textcolor{red}{2}/\textcolor{red}{0.99} \\ \hline

OFFICE & 6000 & 4/1.15 & \textcolor{blue}{3}/0.92 & \textcolor{blue}{3}/\textcolor{blue}{0.8} & \textcolor{blue}{3}/1.15 & \textcolor{blue}{3}/\textcolor{blue}{0.8} & \textcolor{red}{2.4}/\textcolor{red}{0.69} & 30 & \textcolor{blue}{5}/\textcolor{blue}{1.47} & 9/2.21 & 9/2.29 & 7/1.96 & \textcolor{blue}{5}/\textcolor{blue}{1.47} & \textcolor{red}{4}/\textcolor{red}{1.13} \\ \hline

PUMPKIN & 4000 & 7/1.69 & 5/1.30 & \textcolor{red}{4}/\textcolor{red}{1.0} & \textcolor{red}{4}/1.34 & \textcolor{red}{4}/\textcolor{blue}{1.1}  & \textcolor{red}{4.0}/\textcolor{red}{1.03}& 20 & \textcolor{blue}{8}/\textcolor{blue}{1.70} & \textcolor{red}{7}/\textcolor{red}{1.68} & \textcolor{blue}{8}/1.96 & \textcolor{red}{7}/1.77 & \textcolor{red}{7}/1.75 & \textcolor{red}{7}/1.85 \\ \hline

REDKITCHEN & 7000 & 5/1.72 & \textcolor{blue}{4}/1.40 & \textcolor{blue}{4}/\textcolor{blue}{1.2} & \textcolor{blue}{4}/1.68 & \textcolor{blue}{4}/1.3 & \textcolor{red}{3.4}/\textcolor{red}{1.13} & 35 & 7/\textcolor{blue}{1.89} & 7/2.02 & 10/2.63 & 8/2.19 & \textcolor{blue}{6}/1.93 & \textcolor{red}{5}/\textcolor{red}{1.64} \\ \hline

STAIRS & 2000 & \textcolor{blue}{4}/\textcolor{blue}{1.01} & 5/1.47 & \textcolor{red}{3}/\textcolor{red}{0.8} & \textcolor{red}{3}/1.16 & \textcolor{blue}{4}/1.1  & \textcolor{red}{3.2}/\textcolor{red}{0.81}& 20 & 10/2.21 & 18/4.8 & 13/4.24 & 120/27.37 & \textcolor{red}{5}/\textcolor{red}{1.47} & \textcolor{blue}{7}/\textcolor{blue}{1.85} \\ \bottomrule
\end{tabular}}
\label{7scenes_table}
\end{table*}

\begin{table*}[!t]
\centering
\caption{Quantitative comparison of methods on the 7-Scenes dataset with SfM ground truth. Results: MS-Transf \cite{shavit2021learning}, Marepo \cite{chen2024map}, DFNet \cite{chen2022dfnet}, DSAC* \cite{brachmann2021visual}, ACE \cite{brachmann2023accelerated}, GLACE \cite{wang2024glace}, MCLoc \cite{trivigno2024unreasonable}, NeFeS \cite{chen2024neural}, NeRFMatch \cite{zhou2024nerfect}.}
\renewcommand{\arraystretch}{1.5}
\resizebox{\textwidth}{!}{
\begin{tabular}{l|c|c|c|c|c|c|c|c|c|c|c||c|c}
\toprule
\multirow{1}{*}{\makecell{Methods }} & \multirow{2}{*}{\#Images} & \multicolumn{3}{c|}{Absolute pose regression}& \multicolumn{3}{c|}{Scene coordinate regression} & \multicolumn{4}{c||}{Analysis-by-synthesis}& \multirow{2}{*}{\#Images}& \multirow{2}{*}{Ours}\\ \cline{3-12}

 (SfM GT)&  & MS-Transf & Marepo  & DFNet & DSAC* & ACE & GLACE & MCLoc & NeFeS & NeRFMatch & Ours & & \\ \hline
CHESS & 4000 &11/6.4 &1.9/0.83 & 3/1.1 &  \textcolor{blue}{0.5}/\textcolor{blue}{0.17}& \textcolor{blue}{0.5}/0.18& 0.6/0.18& 2/0.8& 2/0.8& 0.9/0.3& \textcolor{red}{0.4}/\textcolor{red}{0.10}& 20& 0.5/0.16\\ \hline

FIRE & 2000 &23/11.5 & 2.3/0.92& 6/2.3& \textcolor{blue}{0.8}/\textcolor{blue}{0.28}& \textcolor{blue}{0.8}/0.33& 0.9/0.34& 3/1.4& 2/0.8& 1.1/0.4& \textcolor{red}{0.6}/\textcolor{red}{0.18}& 10& 0.8/0.26\\ \hline

HEADS & 1000 & 13/13.0& 2.1/1.24& 4/2.3& \textcolor{red}{0.5}/0.34& \textcolor{red}{0.5}/\textcolor{blue}{0.33}& \textcolor{blue}{0.6}/0.34& 3/1.3& 2/1.4& 1.5/1.0& \textcolor{red}{0.5}/\textcolor{red}{0.26}& 10& 0.7/0.48\\ \hline

OFFICE & 6000 &18/8.1 & 2.9/0.93& 6/1.5& 1.2/0.34& \textcolor{blue}{1}/\textcolor{blue}{0.29}& 1.1/\textcolor{blue}{0.29}& 4/1.3& 2/0.6& 3.0/0.8& \textcolor{red}{0.7}/\textcolor{red}{0.22}& 30& 1.2/0.34\\ \hline

PUMPKIN & 4000 &17/8.4 & 2.5/0.88& 7/1.9& 1.2/\textcolor{blue}{0.28}& 1.2/\textcolor{blue}{0.28}& \textcolor{blue}{1}/\textcolor{red}{0.22}& 5/1.6& 2/0.6& 2.2/0.6& \textcolor{red}{0.7}/\textcolor{red}{0.22}& 20& 1.1/1.29\\ \hline

REDKITCHEN & 7000 & 16/8.9& 2.9/0.98& 7/1.7& \textcolor{blue}{0.7}/0.21 & 0.8/\textcolor{blue}{0.20}& 0.8/\textcolor{blue}{0.20}& 6/1.6 & 2/0.6& 1.0/0.3& \textcolor{red}{0.5}/\textcolor{red}{0.14}& 35& 0.9/.022\\ \hline

STAIRS & 2000 & 29/10.3& 5.9/1.48& 12/2.6& \textcolor{blue}{2.7}/\textcolor{blue}{0.78} & 2.9/0.81& 3.2/0.93& 6/2.0 & 5/1.3& 10.1/1.7 & \textcolor{red}{1.6}/\textcolor{red}{0.43}& 20& 4.1/1.10\\ \bottomrule
\end{tabular}}
\label{7scenes_table2}
\end{table*}

\setlength{\textfloatsep}{0pt}  
\setlength{\abovecaptionskip}{0pt} 
\setlength{\belowcaptionskip}{0pt} 

\begin{table}[!t]
\centering
\caption{Quantitative comparison of methods on LLFF and Mip-NeRF 360.}
\renewcommand{\arraystretch}{2} 
\resizebox{\columnwidth}{!}{
\begin{tabular}{l|c|c|c|c|c}
\toprule
\multicolumn{1}{l|}{\shortstack{Methods \\ ($<$0.05 unit/$<$5°)}} & iNerf ($\delta_{s}$) & iComMa (\(\delta_{s}\)) & iComMa (\(\delta_{m}\)) & Ours & Ours (few-shot) \\ \midrule
LLFF & 94.8/72.2 & \textcolor{blue}{99.1}/\textcolor{blue}{99.3} & 75.4/98.2 & \textcolor{red}{100}/\textcolor{red}{100} & \textcolor{red}{100}/\textcolor{red}{100} \\ \hline
Mip-NeRF 360 & 85.6/79.6 & 86.7/90.6 & 68.8/74.8 & \textcolor{red}{100}/\textcolor{red}{100}  & \textcolor{blue}{94.7}/\textcolor{blue}{99.9}\\
\bottomrule
\end{tabular}}
\label{llff_mip_to}
\end{table}

\setlength{\textfloatsep}{0pt}  
\setlength{\abovecaptionskip}{0pt} 
\setlength{\belowcaptionskip}{0pt} 

\subsection{Implementation Details}
Each scene is iterated 30,000 times during GS map construction. Every 20 iterations, a pseudo view is randomly selected to add additional regularization. The weight $\lambda_{d}$ of $\mathcal{L}_{d}$ is 0.05 and the weight $\lambda_{reg}$ of $\mathcal{L}_{reg}$ is 0.01. For localization, we choose the Adam optimizer for gradient descent. The learning rates, including angular, translational, and brightness parameters, are all set to 0.01. The three thresholds for $M_\text{grad}$, $M_\text{fea}$, and $M_\text{occ}$ are set as $1$, $10$, and $0.99$ respectively. The weight $\lambda_{geo}$ for depth residual $E_{geo}(p)$ is set as 0.01. We train and evaluate all datasets on one RTX 4080 Ti GPU with a memory of 16GB. 

\subsection{Comparison}
\textbf{Mip-NeRF 360 and LLFF:} TABLE \ref{llff_mip_to} shows the success rates of iNeRF, iComMa, and LoGS in the LLFF and Mip-NeRF 360 datasets. iNeRF and iComMa depends heavily on pose initialization. $\delta_{s}$ corresponds to a minimal margin initialization where the translation is randomly set from $\pm[0, 0.1]$ in units and the rotation from $\pm[0, 20]$ in degrees. $\delta_{m}$ corresponds to a middle margin initialization where the translation is randomly set from $\pm[0.1, 0.2]$ and the rotation from $\pm[20, 40]$. We first follow the same split setting as iNeRF and iComMa, where most images are used for map construction while only five are used for localization. LoGS perfectly solved this localization problem when tested on five images, achieving a 100\% recall rate with rotation errors less than 5 degrees and translation errors under 0.05 units. 

Discovering this, we further explored a much more difficult few-shot setting, using the Mip-NeRF 360 dataset by uniformly selecting one-tenth of the data from each scene for training (from 12 to 31 images), with the remaining data reserved for testing. For the LLFF dataset, one-fifth of the data was used for training (from 4 to 12 images). Even with such scarce posed images, LoGS reaches higher success rates than the other two methods, demonstrating an advanced competence for accurate pose estimation. Our success on these two datasets is partially due to the new training loss, which significantly improves the rendering quality of the GS map.\\

\begin{figure*}[!t]
    \centering
    \includegraphics[width=0.99\textwidth]{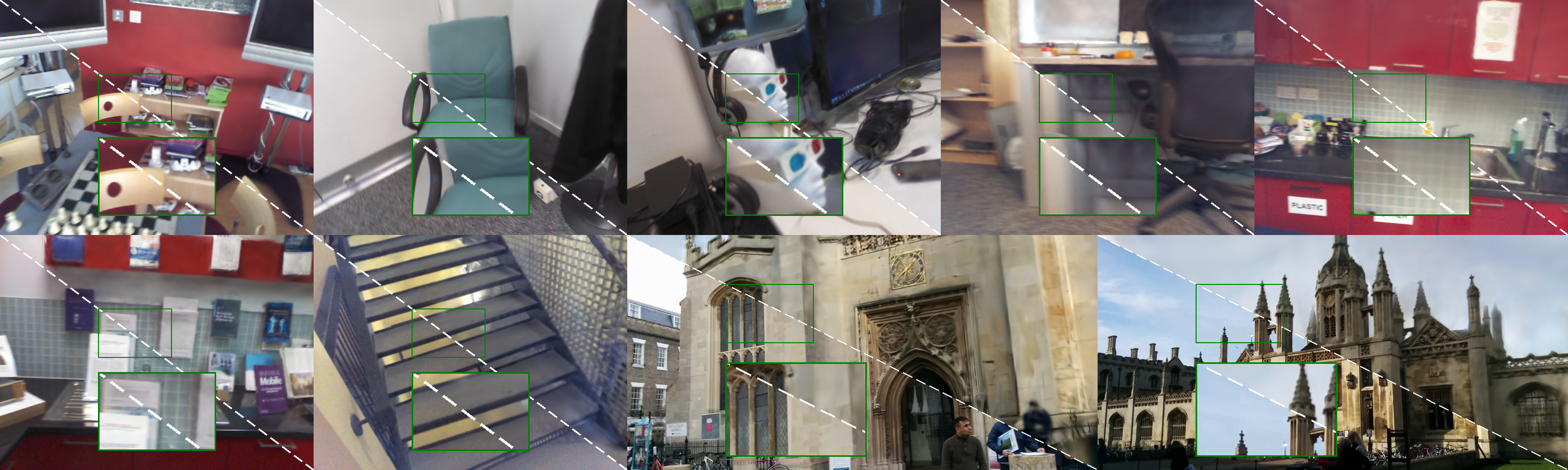}
\caption{Median error pose illustration (full-training on SfM ground truth). The bottom-left region of each plot is the original image. The upper-right part corresponds to the rendered image from Gaussian Splatting and the estimated pose. The first 7 plots are from the 7-scenes datasets and the last two are from the Cambridge Landmarks dataset.}
\label{7_render}
\end{figure*}
\setlength{\textfloatsep}{5pt}  
\setlength{\abovecaptionskip}{5pt} 
\setlength{\belowcaptionskip}{0pt} 

\begin{table*}[!t]
\centering
\caption{Quantitative comparison of methods on the Cambridge Landmarks dataset. SCRNet refers to \cite{li2020hierarchical}.}
\renewcommand{\arraystretch}{2}
\resizebox{\textwidth}{!}{ 
\begin{tabular}{l|c|c|c|c|c|c|c||c|c|c|c|c|c|c}
\toprule
\multirow{2}{*}{Methods} & \multirow{2}{*}{\# Images} & \multicolumn{6}{c||}{Original training (median pose error in cm/°)} & \multirow{2}{*}{\# Images} & \multicolumn{5}{c}{Few-shot training (median pose error in cm/°)} \\ \cline{3-8} \cline{10-15}
 &  & AS & HLoc & SCRNet & HSCNet & DSAC* & Ours& & HLoc  & DSAC* & HSCNet & SP+Reg & FSRC & Ours \\ \hline
GREATCOURT & 1531 & 24/0.13 & \textcolor{blue}{16}/\textcolor{blue}{0.11} & 125/0.6 & 28/0.2 & 49/0.3 & \textcolor{red}{12.7}/\textcolor{red}{0.09} &16 & \textcolor{blue}{72}/\textcolor{blue}{0.27} & NA & NA & NA & 81/0.47 & \textcolor{red}{68}/\textcolor{red}{0.20}\\ \hline

KINGS-COLLEGE & 1220 & 13/0.22 & \textcolor{blue}{12}/\textcolor{blue}{0.20} & 21/0.3 & 18/0.3 & 15/0.3 & \textcolor{red}{10.8}/\textcolor{red}{0.19} & 13 & \textcolor{blue}{30}/\textcolor{blue}{0.38} & 156/2.09 & 47/0.74 & 111/1.77 & 39/0.69 & \textcolor{red}{24}/\textcolor{red}{0.33}\\ \hline

OLDHOSPITAL & 895 & 20/0.36 & \textcolor{red}{15}/\textcolor{red}{0.30} & 21/\textcolor{blue}{0.3} & 19/\textcolor{blue}{0.3} & 21/0.4 & \textcolor{red}{14.6}/\textcolor{blue}{0.31} & 9 & \textcolor{red}{28}/ \textcolor{blue}{0.42} & 135/2.21 & \textcolor{blue}{34}/\textcolor{red}{0.41} & 116/2.55 & 38/0.54 & \textcolor{red}{28}/0.43\\ \hline

SHOPFACADE & 229 & \textcolor{red}{4}/\textcolor{blue}{0.21} & \textcolor{red}{4}/\textcolor{blue}{0.20} & 6/0.3 & 6/0.3 & 5/0.3 & \textcolor{red}{4.1}/\textcolor{red}{0.19} & 3 & 27/1.75 & NA & \textcolor{blue}{22}/\textcolor{blue}{1.27} & NA & \textcolor{red}{19}/\textcolor{red}{0.99} & 39/2.39\\ \hline

STMARYSCHURCH & 1487 & 8/0.25 & \textcolor{red}{7}/\textcolor{blue}{0.21} & 16/0.5 & 9/0.3 & 13/0.4 & \textcolor{red}{6.9}/\textcolor{red}{0.20} & 15 & 25/0.76 & NA & 292/8.89 & NA & 31/1.03 & \textcolor{red}{22}/\textcolor{red}{0.67}\\ \bottomrule
\end{tabular}}
\label{cambridge_table_fewshot}
\end{table*}
\setlength{\textfloatsep}{0pt}  
\setlength{\abovecaptionskip}{0pt} 
\setlength{\belowcaptionskip}{0pt} 

\textbf{7-scenes:} Each cell of Table \ref{7scenes_table} contains the median translation error (in centimeter) and the median rotation error (in degree), respectively. The left side of the table shows the localization accuracy obtained by each approach being trained on the full training set, while the right side shows the accuracy of the few-shot training sets. As the training data decreases, the localization error increases for all methods. The ability to achieve accurate localization under such extreme conditions demonstrates the stability of a system.

With all the data, LoGS achieved the best results across seven scenes. When using only a handful of images, it outperforms other methods in multiple scenes, with the median rotational error in the PUMPKIN scene being nearly identical to the best result, while the translational and rotational errors in the STAIRS scene show a relative gap compared to FSRC \cite{dong2022visual}. Upon analysis, we believe this is due to 1) the similarly colored, repetitive structure of the multi-layered steps in the stairs and 2) the significant deviation in the initial pose estimation, which together cause the model to converge to a local optimum. 

We also train on SfM ground truth and obtain the median error results for all 7 scenes (see TABLE \ref{7scenes_table2}). Brachmann et al.~\cite{brachmann2021limits} suggest no significant advantage of one ground truth over the other on the 7-Scenes dataset. However, different localization methods show varying accuracy depending on the type used. Moreover, NeRF-synthesis methods \cite{chen2024neural,zhou2024nerfect} have demonstrated that rendered images tend to have higher quality when using SfM ground truth, and we observed the same phenomenon with the GS map. LoGS sets a new baseline for analysis-by-synthesis approaches trained with whole data. Utilizing only a few dozen images, We found that LoGS achieve median translation error around a centimeter (except the STAIRS scene). This is a remarkably impressive result, as the achieved accuracy is comparable to SCR methods trained with one hundred times more data.

\setlength{\parskip}{8pt}
\textbf{Cambridge Landmarks:}
TABLE \ref{cambridge_table_fewshot} summarizes the median pose errors in centimeter and degree. LoGS, in general, demonstrates accuracy improvements over state-of-the-art feature matching-based methods on whole dataset training. We then test LoGS with around 1\% data. NA indicates failure: median translation error greater than 500 centimeter. First, it is worth noting that many methods using neural networks as map frameworks, such as DSAC*, failed. This is because these methods employ complex network structures to enhance learning capability, which leads to poor generalization with a small training set. Nevertheless, we achieved the best accuracy in four scenes, setting a new benchmark. Overall, LoGS demonstrated robustness in large-scale outdoor scenes with limited training data. Our "failure" on the SHOPFACADE scene is mainly because it is a corner, and three simple RGB images made it difficult for 3DGS to determine depth, resulting in a final map with a few overlapping shadows.

\setlength{\parskip}{10pt}
\section{Conclusion}
\setlength{\parskip}{0pt}
This paper broadens the boundaries of mobile robotics~\cite{kanoulas2019curved, kanoulas2018rxkinfu, jiao2024RTMapping, jiao2024LiteVLoc} by exploring visual localization using 3DGS as a map representation. Scene Coordinate Regression and Absolute Pose Regression can accurately estimate poses with abundant posed images but tend to fail when training viewpoints are insufficient. In contrast, feature-based methods can predict poses under both conditions but with less accuracy. Our pipeline LoGS achieved high-precision image rendering from the GS map by optimizing the initial point cloud, loss function, and regularization methods. Based on that, LoGS combined multiple masks, selected the most representative pixels to compare photometric loss on RGB(D) channels, and utilized gradient descent to obtain an accurate pose from an initial estimation. Our method outperformed baselines in full/few-shot settings on four large-scale datasets, achieving leading-edge results. Future directions on finer GS reconstruction (e.g., illumination changes), new masking strategies, and GS map compression that reduces memory and increases localization speed can improve the work.

\newpage

\bibliographystyle{unsrt}
\bibliography{references}

\end{document}